\documentclass[runningheads]{llncs}
\usepackage{graphicx}
\usepackage{tikz}
\usepackage{comment}
\usepackage{amsmath,amssymb} % define this before the line numbering.
\usepackage{color}
\usepackage{orcidlink} 
\usepackage[accsupp]{axessibility}  % Improves PDF readability for those with disabilities.

\usepackage{floatrow}
\usepackage{multirow}
\usepackage{array}
\usepackage{booktabs}

\begin{document}
% \linenumbers
\pagestyle{headings}
\mainmatter
\def\ECCVSubNumber{1016}  % Insert your submission number here
\title{Dual-Stream Knowledge-Preserving Hashing for Unsupervised Video Retrieval} % Replace with your title
%With a little help from my friend layer: 
% \begin{comment}
% \titlerunning{ECCV-22 submission ID \ECCVSubNumber} 
% \authorrunning{ECCV-22 submission ID \ECCVSubNumber} 
% \author{Anonymous ECCV submission}
% \institute{Paper ID \ECCVSubNumber}
% \end{comment}
% ******************

% \begin{comment}
\titlerunning{Dual-Stream Knowledge-Preserving Hashing}
% an abbreviated paper title here 0000-0002-6249-5315
%
\author{Pandeng Li\inst{1}\orcidlink{0000-0002-0717-8659} \and
Hongtao Xie\inst{1}\orcidlink{0000-0002-6249-5315} \thanks{H. Xie is the corresponding author.} \and
% Hongtao Xie\inst{1} \thanks{H. Xie is the corresponding author. The results and code are available at \url{https://github.com/IMCCretrieval/DKPH.}} \and
Jiannan Ge\inst{1}\orcidlink{0000-0002-2580-9055} \and
Lei Zhang\inst{2}\orcidlink{0000-0002-2839-8693}  \and
Shaobo Min \inst{3}\orcidlink{0000-0002-7700-2149} \and
Yongdong Zhang \inst{1}\orcidlink{0000-0002-1151-1792} }
\authorrunning{Li et al.}
\institute{University of Science and Technology of China \\ 
\and
Kuaishou Technology   \and
Tencent Data Platform\\
\email{\{lpd,gejn\}@mail.ustc.edu.cn,} \email{\{htxie,zhyd73\}@ustc.edu.cn,} \email{zhanglei06@kuaishou.com,}
\email{bobmin@tencent.com}}
% \end{comment}
%******************
\maketitle

\begin{abstract}
% Such reconstruction  constraint spends much effort on dynamic temporal changes without focusing on stationary global semantics that are more useful for retrieval.   
%显示的捕捉上面两个信息
Unsupervised video hashing usually optimizes binary codes by learning to reconstruct input videos.
Such reconstruction constraint spends much effort on frame-level temporal context changes without focusing on video-level global semantics that are more useful for retrieval.
Hence, we address this problem by decomposing video information into reconstruction-dependent and semantic-dependent information, which disentangles the semantic extraction from reconstruction constraint.
Specifically, we first design a simple dual-stream structure, including a temporal layer and a hash layer.
Then, with the help of semantic similarity knowledge obtained from self-supervision, the hash layer learns to capture information for semantic retrieval, while the temporal layer learns to capture the information for reconstruction.
In this way, the model naturally preserves the disentangled semantics into binary codes.
Validated by comprehensive experiments, our method consistently outperforms  the state-of-the-arts on three video benchmarks.


\end{abstract}

\section{Introduction}
In view of the explosive growth of informative media (\textit{i.e.,} videos)~\cite{zhang2020graph,liu2018temporal,contournet,wu2022end,min2020domain},
 the efficient large-scale retrieval system~\cite{brown2020smooth,NASA,yu2021heterogeneous,ge2021semantic,DFRQ,PQN} has become an urgent requirement in the real world. 
Video retrieval system needs to understand the semantic similarity information implicit in videos~\cite{zhang2018cross}, which can be found by comparing the real-valued features in the last layer of deep networks.
Unfortunately, these massive amounts of features take up large storage space~\cite{cui2020exchnet} and seriously affect the retrieval speed.
As a key building block of search algorithms, hashing~\cite{ITQ}, can alleviate the above issue by compressing high dimensional features into compact binary codes.
However, the abundant content and temporal dynamics of videos make it difficult for binary codes to preserve the similarity structure of the  real-valued feature space~\cite{liong2016deep,UDVH}.
Besides, compared to image datasets, the manual annotation and pre-training costs of standard large-scale video data are very high~\cite{MMT}.
Therefore, unsupervised video hashing has intrigued many researchers in practice~\cite{SSVH,NPH,BTH,SSTH}.

\begin{figure}[t]
	\centering
	\includegraphics[width= 0.9\linewidth]{ 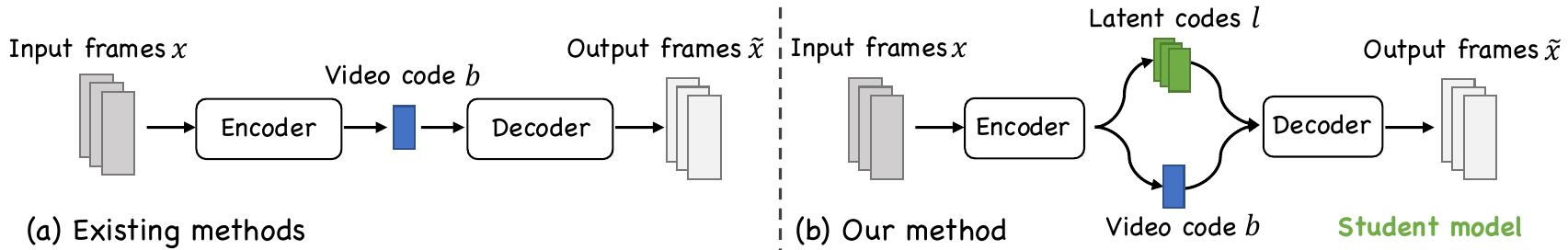}
	\caption{  (a) Existing methods usually optimize binary codes by using an encoder-decoder architecture to reconstruct the visual information of input frames. (b) Our method utilizes latent codes in the student model  to model temporal changes required for the reconstruction task, thereby allowing binary codes to focus on global semantics. }
	\label{fig:exist}
\end{figure}

As shown in Fig.~\ref{fig:exist} (a), existing unsupervised video hashing methods usually optimize binary codes by using an encoder-decoder architecture to reconstruct the visual information of input frames.
For example, Zhang \textit{et al.}~\cite{SSTH} employ an encoder-decoder Recurrent Neural Networks (RNNs)~\cite{RNN} to capture the temporal nature of videos for binary codes.
Later, Li \textit{et al.}~\cite{UVVH} introduce Variational Auto-Encoders (VAE)~\cite{VAE} to learn a probabilistic latent code of video variations.
% Song~\textit{et al.}~\cite{SSVH} use the Long Short Term Memory (LSTM) networks~\cite{srivastava2015unsupervised} to model more granular inter-frame dependencies in videos.
% The binary codes of these methods are forced to meet the goal of data reconstruction.
% where binary codes are forced to satisfy the goal of data reconstruction.
However, these binary codes are forced to independently satisfy the goal of video reconstruction, which may be sub-optimal for semantic retrieval due to the heterogeneity of two tasks~\cite{guo2018dynamic} (\textit{i.e.,} the retrieval and reconstruction tasks).

\begin{figure}[t]
	\centering
	\includegraphics[width= 0.85 \linewidth]{ 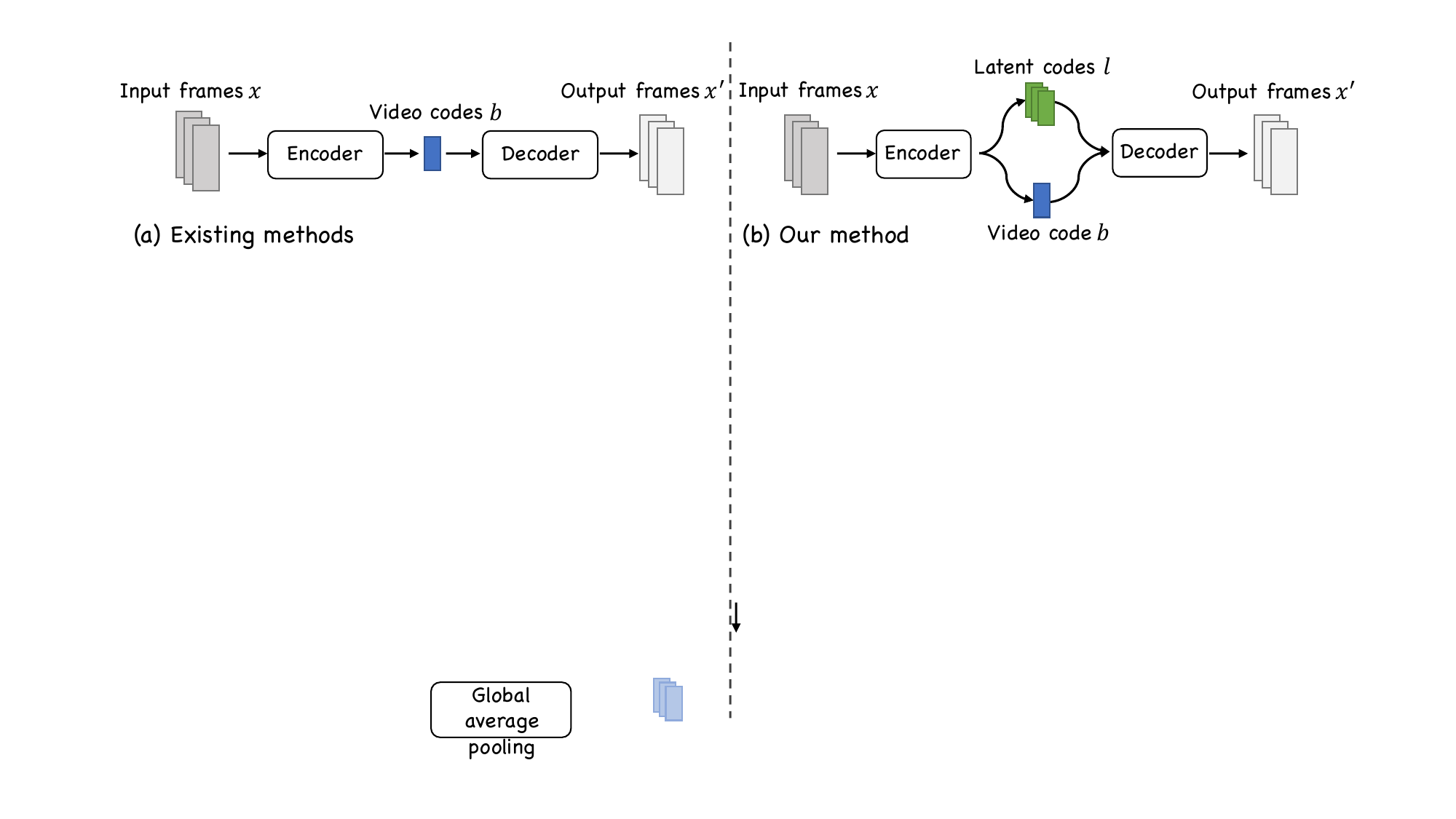}
	\caption{Essential and superfluous information for the semantic video retrieval task. }
	\label{fig:information}
\end{figure}

%Given the original video as input, existing models tend to compress information that is good for reconstruction but bad for semantic discrimination.
%given raw videos as input, hash model tends to compress the information that is conductive to reconstruction but not to semantic identification with the construction task.

% Specifically, given raw videos as input, deep networks tend to compress task-essential information while removing task-superfluous information.
Specifically, given raw videos as input, existing models tend to compress the information that is essential to reconstruction but may be superfluous for similarity search.
This argument can be proved in Information Bottleneck (IB) ~\cite{IB} from an information-theoretic perspective.
IB models the information flow~\cite{CDVAE} from input  $x$ to  the target  $\tilde{x}$ through latent variable $b$ (\textit{e.g.,} binary codes), where the optimal $b$ should contain the minimal sufficient information to predict $\tilde{x}$ but discards all superfluous information in $x$ that is irrelevant for $\tilde{x}$.
This provides an optimization principle that maximizes the mutual information $I(b; \tilde{x})$ between the latent variable and the target, and simultaneously constrains $I(x; b)$ small.
In the above existing hashing methods, maximizing $I(b; \tilde{x})$ corresponds to minimizing the reconstruction error. 
The reconstruction-essential information~\cite{CIBHash} may be the sequence of actions, constant changes, etc., which involves more fine-grained temporal understanding.
% Thus, binary codes spend much effort on dynamic temporal changes, instead of focusing on learning stationary global semantics.
% concepts.
However, as shown in Fig.~\ref{fig:information}, binary codes require more attention to global semantic concepts like ``biking'' or ``cat'' for ranking videos.
% distinguishing videos.
Because, the retrieval goal is not to retain all information of the original video data, but to preserve the discriminative similarity information.
% for better retrieval.

% decomposing video information with teacher-student optimization to obtain semantic binary codes.

Based on the above discussion, we propose a novel Dual-stream Knowledge-Preserving Hashing (DKPH) framework to obtain semantic binary codes by decomposing video information into semantic-dependent and reconstruction-dependent information.
% to preserve the disentangled semantics into binary codes.
% , which disentangles the semantic extraction from reconstruction constraint.
As shown in Fig.~\ref{fig:framework}, DKPH fully releases the potential of semantic learning via teacher-student optimization:
(1) the student model designs a simple but effective dual-stream structure to disentangle the semantic extraction from reconstruction constraint on a single binary code;
% relieve the pressure of a single binary code facing reconstruction constraints;
(2) the teacher model refines the semantic similarity knowledge to further guide the meaningful information decomposition in the student model.
% , thereby generating discriminative semantic binary codes in the student model.
% divide the information flow inside layers and decompose video information within features.

More concretely, the dual-stream structure contains a parallel temporal layer and hash layer.
% that decompose video information into reconstruction-dependent and semantic-dependent parts. 
The temporal layer tries to capture reconstruction-dependent information by learning dynamic frame-level features, 
% which relieves the reconstruction pressure of binary codes.
while the hash layer focuses on the semantic-dependent part from a global video-level perspective.
% output a  semantic-dependent discriminative binary code from a global video-level perspective, guided by the semantic similarity knowledge from the teacher model.
% To further guide the information decomposition, 
To achieve the above goal, a teacher model is trained in a self-supervised manner to construct a Gaussian-adaptive similarity graph, which captures the inherent similarity relations between samples.
This relation knowledge is preserved into the student hash layer to generate semantic-dependent discriminative binary codes.
% via, which is preserved into the student hash layer to guide the information decomposition.
% More concretely, inspired by the visual cloze task in~\cite{bert,BTH}, we first train the teacher model to initially obtain compact visual embeddings.
% During this guidance process (\textit{i.e.,} knowledge preservation), we first use the trained transformer in the teacher model~\cite{BTH} to initially obtain compact visual embeddings.
% These embeddings capture long-term inter-frame correlations and reduce the subsequent computational complexity due to low embedding dimensionality.
% Based on the semantic knowledge from visual embeddings, we construct a Gaussian-adaptive similarity graph that captures the inherent sample relations.
% by estimating positives and hard negatives of training videos.
% Finally, these relations can guide the student hash layer to generate semantic-dependent discriminative binary codes and maintain the neighborhood structure in Hamming space.

\noindent \textbf{Contributions.}
(1) We propose a novel framework, DKPH, to fully release the potential of semantic learning on binary codes and may shed critical insights for the retrieval community.
To our best knowledge, our method is the first work that explores the task heterogeneity in video hashing. 
(2) A simple but effective dual-stream structure is developed  to decompose video information, which can generate  semantic-dependent discriminative binary codes by preserving the semantic similarity knowledge from the proposed Gaussian-adaptive similarity graph.
(3) Extensive experiments demonstrate that DKPH outperforms  state-of-the-art video hashing models on FCVID, ActivityNet and YFCC datasets.

% the video retrieval task using 

%  https://arxiv.org/abs/physics/0004057

\section{Related Work}
\noindent \textbf{Unsupervised hashing.}
Unsupervised hashing aims to learn hash functions 
that compress data points into binary codes, which are built on training data without manual annotations.
Iterative quantization (ITQ)~\cite{ITQ} is a traditional representative method that directly explores the minimum quantization error by learning an optimal rotation of principal component directions.
However, non-deep image hashing methods only seek a single linear projection, resulting in poor generalization.
Then, Deep Hashing (DH)~\cite{DH} uses a deep neural network
to learn binary codes via multiple hierarchical non-linear transformations.

Due to the explosive growth of short videos, some works~\cite{MFH,NPH} also focus on video hashing.
Multiple Feature Hashing (MFH)~\cite{MFH} mines local structural information while ignoring inter-frame temporal consistency~\cite{VHDT}.
Later, a series of methods based on encoder-decoder structure have become mainstream methods for video hashing. 
For example, Self-Supervised Temporal Hashing (SSTH)~\cite{SSTH} employs an encoder-decoder RNNs to capture the temporal nature of videos.
Li~\textit{et al.}~\cite{JTAE} jointly model static visual appearance and temporal pattern into binary codes via two special reconstruction losses. Unsupervised Deep Video Hashing (UDVH)~\cite{UDVH} emphasizes balancing dimensional variation for each binary representation.
Self Supervised Video Hashing (SSVH)~\cite{SSVH} attempts more powerful Bi-LSTM to model more granular inter-frame dependencies.
Despite a similar network architecture to SSVH, Neighborhood Preserving Hashing (NPH)~\cite{NPH} encodes the neighborhood-dependent video content as a binary code.
Bidirectional Transformer Hashing (BTH)~\cite{BTH} introduces the BERT architecture~\cite{bert} in NLP to explore inter-frame correlations, and achieves excellent results.
However, these video hashing methods fail to consider the heterogeneity between reconstruction and retrieval tasks for optimizing binary codes.
Recently, Shen~\textit{et al.}~\cite{shen2020auto} propose twin bottlenecks to extract continuous features, but the similarity optimization process for binary codes is still implicit and heavily depends on the reconstruction effects.
 Besides, more efficient sample relations have not been explored, which affects the semantic discriminative of binary codes.

\noindent \textbf{Knowledge distillation.}
\cite{hinton2015distilling,romero2014fitnets} first propose to transfer knowledge from teacher models to student models through the soft outputs or intermediate layer features.
Recently, Knowledge Distillation (KD) is extended to training deep 
networks in generations and \cite{bagherinezhad2018label,liu2019semantic} find that KD can refine ground truth labels. 
In unsupervised video hashing, to preserve and distill the semantic knowledge, we refine pre-trained CNN features to visual embeddings in the teacher model, which can further construct an efficient similarity graph for training student model. 

\section{Method}
\subsection{Problem Definition}
We introduce some notations and the problem definition of unsupervised video hashing.
Generally, learning hash functions is considered in an unsupervised manner from a training set of $N$ video data points $\mathcal{V} = \{\boldsymbol{v}_{i}\}^{N}_{i=1} \in \mathbb{R}^{N \times M \times D} $, where each $\boldsymbol{v}_{i} = [\boldsymbol{x}_1, \cdots,\boldsymbol{x}_M] \in \mathbb{R}^{M \times D} $ is a CNN feature set, $M$ is the number of frames, and $D$ is the feature dimension of each frame.
DKPH aims to learn nonlinear hash functions based on transformer blocks that map each video data point $\boldsymbol{v}_{i}$ into a $K$-dimensional Hamming space $\boldsymbol{b}_{i} \in  \{-1,1\}^{K} $, which needs to keep relative semantic similarity between videos.

% % produce semantic-dependent  binary codes for efficient video search.
% Generally, the hash function $H: \mathbb{R}^{M \times D} \rightarrow \{-1,1\}^{K}  $ maps $\boldsymbol{v}_{i}$ into a $K$-dimensional Hamming space $\boldsymbol{b}_{i} = H (\boldsymbol{v}_{i})$, which needs to keep relative semantic similarity between videos.

\subsection{Network Overview}
\begin{figure}[t]
	\centering
	\includegraphics[width=0.95 \linewidth]{ 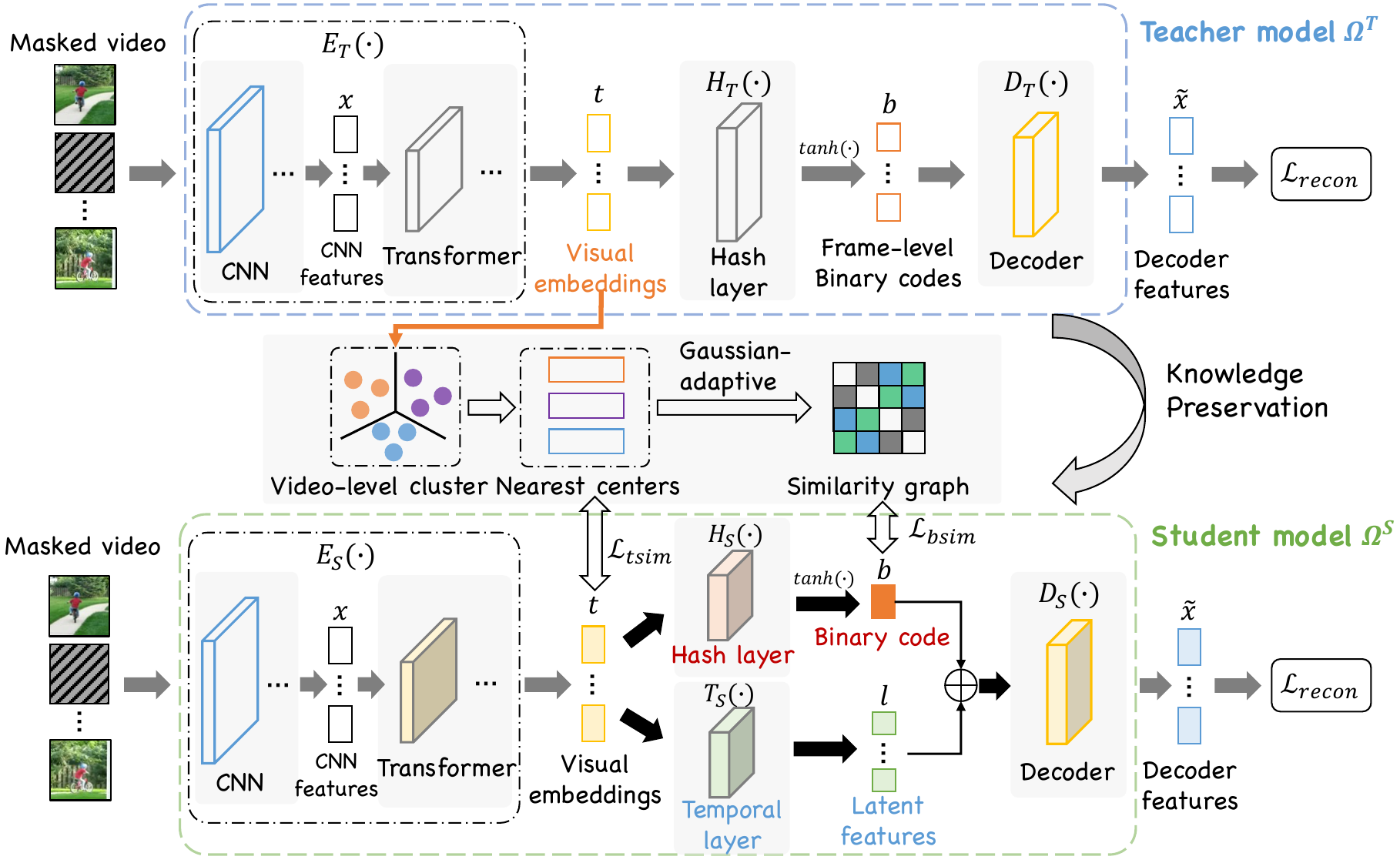}
	\caption{The proposed DKPH framework which involves
	 (1) training teacher model $\boldsymbol{\Omega}^{T}$ and student model $\boldsymbol{\Omega}^{S}$ in an unsupervised manner,
	 (2) distilling semantic knowledge from $\boldsymbol{\Omega}^{T}$ to guide the information decomposition in $\boldsymbol{\Omega}^{S}$. }
	\label{fig:framework}
\end{figure}

DKPH consists of a teacher model
$\boldsymbol{\Omega}^{T}$ and a student model $\boldsymbol{\Omega}^{S}$.
As shown in Fig.~\ref{fig:framework}, $\boldsymbol{\Omega}^{T}$ is a common encoder-decoder architecture that can exchange inter-frame information through transformers to obtain long-term semantic knowledge.
 $\boldsymbol{\Omega}^{S}$ is a dual-stream encoder-decoder architecture that can disentangle the semantic extraction and reconstruction constraint on a single binary code to better capture the semantic information transmitted by $\boldsymbol{\Omega}^{T}$.
%  for video retrieval.
%  relieves the pressure of a single binary code facing reconstruction constraints to 
 In this section, we introduce three key sub-networks:
 transformer encoder, hash layer and temporal layer, where the structure of transformer encoder is the same in  $\boldsymbol{\Omega}^{T}$ and $\boldsymbol{\Omega}^{S}$.
 %在本节中，我们介绍关键的三个子网络：
%Similar to other video hashing methods~\cite{SSTH,JTAE}, we pursue that  videos $\boldsymbol{v}_1, \boldsymbol{v}_2$ sharing the same semantics have binary codes $\boldsymbol{b}_1, \boldsymbol{b}_2$ that are close in Hamming space.

\noindent\textbf{Transformer encoder.}
%Since Transformer has recently demonstrated promising results on many computer vision tasks, 
To model long-term semantic correlation in videos, 
we first employ transformer blocks to handle the pre-processing CNN frame features.
Each transformer encoder block has a multi-head self-attention and a feed-forward layer.
Different from splitting images into several tokens in ViT~\cite{ViT}, we treat frame features $[\boldsymbol{x}_{i}^1, \cdots,\boldsymbol{x}_{i}^M]$ as token units, which contain rich visual content information.
Besides, to learn the ordering information of each frame inside the original video, 
we follow the standard procedure in ViT by adding trainable positional encoding embeddings $\mathbf{E}_{pos}$.
Thus, the input video matrix $\mathbf{X}_i$  is defined as follows:
\begin{equation}
\mathbf{X}_i =[\boldsymbol{x}_{i}^1, \cdots,\boldsymbol{x}_{i}^M] + \mathbf{E}_{pos}.
\end{equation}

Given the input matrix $\mathbf{X}_i$, we calculate  queries $\mathbf{Q}_{i}$, keys $\mathbf{K}_{i}$ and values $\mathbf{V}_{i}$ as follows:
$
\mathbf{Q}_{i}=\mathbf{X}_{i} \mathbf{W}^{Q}_{i}, \quad \mathbf{K}_{i}=\mathbf{X}_{i} \mathbf{W}^{K}_{i}, \quad \mathbf{V}_{i}=\mathbf{X}_{i} \mathbf{W}^{V}_{i}, 
$
where  $\mathbf{W}_{i}^{Q}$, $\mathbf{W}_{i}^{K}$ and $\mathbf{W}_{i}^{V}$ are linear projections with an output of $d$ dimensions. 
Then the self-attention outputs can be calculated by
\begin{equation}
\operatorname{Att}(\mathbf{Q}_{i}, \mathbf{K}_{i}, \mathbf{V}_{i})=\operatorname{softmax}\left(\mathbf{Q}_{i} \mathbf{K}_{i}^{T} / \sqrt{d}\right) \mathbf{V}_{i}^{T}.
\end{equation}
Finally, these frame token units undergo multiple informative interactions, which are transformed into a sequence of visual embeddings $[\boldsymbol{t}_{i}^1, \cdots,\boldsymbol{t}_{i}^M]$.
% These embeddings emphasize the long-term semantic concepts within the video.
% and are beneficial for hash learning.
% which are efficient information preprocessing for hash learning.

\noindent \textbf{Hash layer.}
As shown in the teacher model $\boldsymbol{\Omega}^{T}$ of Fig.~\ref{fig:framework}, the intuitive approach~\cite{BTH} is to directly reduce the visual 
embedding dimension of each frame through linear mapping $[\hat{\boldsymbol{t}}_{i}^1, \cdots,\hat{\boldsymbol{t}}_{i}^M] = {H}_T \left([\boldsymbol{t}_{i}^1, \cdots,\boldsymbol{t}_{i}^M] \right) \in \mathbb{R}^{M \times K}$, 
and then binarize them to obtain frame-level binary codes $[\boldsymbol{b}_{i}^1, \cdots,\boldsymbol{b}_{i}^M] \in \{-1,1\}^{M \times K}$.
However, according to the settings of existing methods~\cite{BTH} in the testing phase, $\boldsymbol{\Omega}^{T}$ needs to average $[\boldsymbol{b}_{i}^1, \cdots,\boldsymbol{b}_{i}^M]$ to obtain a real-valued code, which is binarized to video-level binary code for retrieval.
% perform global average pooling (GAP) on the binary codes of all frames in a vide
This leads to two issues: (1) there is a quantization error  between 
 real-valued codes and binary codes, resulting 
in a sub-optimal solution;
(2) when $+1$ and $-1$ numbers of frame binary codes are the same,
 $\{-1,0,1\}^{K}$ may be generated, which violates the principle of hashing.
%可是，在测试阶段，模型需要将所有帧的二进制码进行全局平均池化，从而得到一个实值码.
%这会导致两个问题：1）实值码和最终的二进制码存在量化误差，导致次优解；
%2）当帧的数目为偶数，帧二进制码的正负1个数相同时，可能会产生三进制码{-1,0,1}

Therefore, we directly concatenate the 
frame visual features $[\boldsymbol{t}_{i}^1, \cdots,\boldsymbol{t}_{i}^M]$ 
 of the video from a global perspective in $\boldsymbol{\Omega}^{S}$, and then 
 extract a real-valued code through the Fully Connected (FC) layer:
\begin{equation}
\hat{\boldsymbol{t}}_{i}= {H}_S \left(Concat([\boldsymbol{t}_{i}^1, \cdots,\boldsymbol{t}_{i}^M])\right) \in \mathbb{R}^K.
\end{equation}
Finally, we can obtain a K-bit binary code:
\begin{equation}
\boldsymbol{b}_{i}=\operatorname{sgn}\left(\tanh(\hat{\boldsymbol{t}}_{i})\right) \in \{-1,1\}^{K}.
\label{binary_code}
\end{equation}
Besides, to avoid the discrete optimization problem~\cite{hashnet}, we  follow~\cite{binarynet} for backpropagating gradients.
In this way, the encoder-decoder methods~\cite{SSTH,SSVH} can compress the visual information as much as possible.
However, to meet the goal of video reconstruction,  the compression process may contain lots of retrieval-superfluous information, which affects the discriminativeness of binary codes.
% while binary codes require more attention to essential semantic concepts.
%In this way,  Our model can encode stationary semantic information from a global video-level perspective

\noindent \textbf{Temporal layer.}
To alleviate the task heterogeneity problem, a simple but effective dual-stream structure is introduced to decompose video information in $\boldsymbol{\Omega}^{S}$.
Specifically, we design a temporal layer $T_S$ parallel to the hash layer $H_S$ in the dual-stream structure.
$T_S$ directly reduces the dimension of frame visual features $[\boldsymbol{t}_{i}^1, \cdots,\boldsymbol{t}_{i}^M]$  to obtain frame-level latent features via FC:
\begin{equation}
[\boldsymbol{l}_{i}^1, \cdots,\boldsymbol{l}_{i}^M]={T}_S  \left([\boldsymbol{t}_{i}^1, \cdots,\boldsymbol{t}_{i}^M]\right) \in \mathbb{R}^{M\times K} .
\label{latent_feature}
\end{equation}
The temporal layer attempts to model complex information such as dynamic temporal changes via the reconstruction constraint, while for the hash layer, we will design similarity constraints to guide the flow of semantic information.
Next, we will introduce how to perform  dual-stream reconstruction and similarity knowledge preservation respectively.

% reconstruction-essential and retrieval-superfluous information all remaining in binary codes,
%现有的视频哈希工作往往设计重建任务想要压缩信息到二进制码中，
\subsection{Dual-stream Reconstruction Learning}
Existing video hashing works usually design the reconstruction task to compress visual information into binary codes.
Inspired by masked language modeling in BERT~\cite{bert,wang2021two}, 
$\boldsymbol{\Omega}^{T}$~\cite{BTH} exploits the visual cloze task to optimize transformer blocks and capture inter-frame 
correlations, which randomly masks the input frame features as tokens and reconstructs the masked tokens in the decoder.
In this way, frame-level binary codes in $\boldsymbol{\Omega}^{T}$  can retain all the essential information for reconstruction, rather than retrieval.

To avoid this issue in $\boldsymbol{\Omega}^{S}$, we first mix $[\boldsymbol{l}_{i}^1, \cdots,\boldsymbol{l}_{i}^M]$ and $\boldsymbol{b}_i$ derived in  Eq.~\ref{binary_code} and Eq.~\ref{latent_feature}, and then leverage the FC layer to reconstruct:
\begin{equation}
[\tilde{\boldsymbol{x}}_{i}^1, \cdots,\tilde{\boldsymbol{x}}_{i}^M]=D_S\left([\boldsymbol{l}_{i}^1 + \boldsymbol{b}_i, \cdots,\boldsymbol{l}_{i}^M + \boldsymbol{b}_i] \right) \in \mathbb{R}^{M\times D} .
\end{equation}
Then, we can use the mean square error loss
 to measure the difference between
  CNN features $[\boldsymbol{x}_{i}^1, \cdots,\boldsymbol{x}_{i}^M]$ and 
  decoder features $[\tilde{\boldsymbol{x}}_{i}^1, \cdots,\tilde{\boldsymbol{x}}_{i}^M]$ in $\boldsymbol{\Omega}^{S}$: 
%In this way, 
\begin{equation}
\mathcal{L}_{recon}=\frac{1}{D N M} \sum_{i=1}^{N} \sum_{m=1}^{M}\left\|\boldsymbol{x}_{i}^{m}-\tilde{\boldsymbol{x}}_{i}^{m}\right\|_{2}^{2}.
\label{eq:recon}
\end{equation}
Please note that, in $\boldsymbol{\Omega}^{T}$, the definition of $\mathcal{L}_{recon}$  is the same as Eq.~\ref{eq:recon}, but teacher decoder features can only be generated from frame-level binary codes $[\boldsymbol{b}_{i}^1, \cdots,\boldsymbol{b}_{i}^M]$,  where the code length is fixed to 128 in experiments.

\subsection{Semantic Knowledge Preservation}
% \vspace{-5pt}
Using the frame-level reconstruction task alone does not make the two layers perform the desired role, so we further guide $H_S$ to learn video-level semantic similarity information.
For unsupervised learning, some image hashing works~\cite{distillhash} prove that neighborhood structures learned from original CNN features can capture the similarity relations between samples.
However, this strategy is time-consuming due to building a similarity graph for all samples directly, and has lots of noisy predictions, which confuses the learning of hash functions.
Benefiting from the teacher-student distillation framework in Fig.~\ref{fig:framework}, 
we construct a Gaussian-adaptive similarity graph from $\boldsymbol{\Omega}^{T}$ that captures the inherent 
semantic relations by estimating positives and hard negatives of training videos.
These relations can guide $H_S$ to generate discriminative binary codes and maintain the neighborhood structure in Hamming space.

Specifically, we first warm up $\boldsymbol{\Omega}^{T}$ with the
reconstruction task
$\mathcal{L}_{recon}$, and exploit the transformer  visual embeddings 
$[\boldsymbol{t}_{i}^1, \cdots,\boldsymbol{t}_{i}^M]$ instead of the high dimensional CNN frame features to mine similarity relations.
Although visual embeddings may contain redundant information due to $\mathcal{L}_{recon}$,
 they model inter-frame correlations compared to CNN features, 
 which are helpful for mining long-term semantic concepts.
To obtain video-level graph, we average $[\boldsymbol{t}_{i}^1, \cdots,\boldsymbol{t}_{i}^M]$ to video embedding $\overline{\boldsymbol{t}}_{i}$.
Then, to solve the time-consuming problem, we follow~\cite{AGH} to use the neighbor graph between each video point $\overline{\boldsymbol{t}}_{i}$ and the cluster center $\{\boldsymbol{c}_{i}\}_{i=1}^{N_c}$ of video points to approximate similarity relations between $\overline{\boldsymbol{t}}_{i}$ and $\{\overline{\boldsymbol{t}}_{i}\}_{i=1}^{N}$, 
where $N_c$ is the number of K-means clustering center.
For each $\overline{\boldsymbol{t}}_{i}$,  we calculate
$p$ nearest centers $\{\boldsymbol{c}_{ij}\}_{j=1}^{p}$, and the similarity matrix $\boldsymbol{Z} \in \mathbb{R}^{N \times N_c}$ is expressed as:
\begin{equation}
\boldsymbol{Z}_{i j}=
\frac{\exp \left(-||\overline{\boldsymbol{t}}_{i}, \boldsymbol{c}_{ij}||_2 / \alpha \right)}{\sum_{j^{\prime}=1}^{p} \exp \left(-||\overline{\boldsymbol{t}}_{i}, \boldsymbol{c}_{i j^{\prime}}||_2 / \alpha\right)}, 
\end{equation}
where $\alpha$ is a bandwidth parameter.
Note that the similarity values between $\overline{\boldsymbol{t}}_{i}$  and corresponding $N_c-p$ non-nearest centers in $\boldsymbol{Z}$ are set to 0, for simplicity.
% and the similarity values in $\boldsymbol{Z}$ between non-nearest neighbor centers and $\overline{\boldsymbol{t}}_{i}$ are set to 0. 
Finally, an approximate graph adjacency $\boldsymbol{A}\in \mathbb{R}^{N \times N} $ is calculated as:
$
\boldsymbol{A} =\boldsymbol{Z} \Lambda^{-1} \boldsymbol{Z}^{T}, 
$
where ${\Lambda}=\operatorname{diag} (\boldsymbol{Z}^{T} \mathbf{1} ) \in \mathbb{R}^{N_c \times N_c}$.
However, $\boldsymbol{A}$ may still be a noisy similarity signal, where
the nearest center number $p$ greatly affects the prediction quality.
To avoid this dilemma, the existing work~\cite{BTH} builds multiple large adjacency matrices to vote for credible sample relations, but it requires careful parameter tuning and takes up huge storage space on large video datasets.
% and is inefficient.
% To avoid this dilemma, existing work~\cite{BTH} builds multiple large adjacency matrices to mine positive and negative samples, but requires careful parameter tuning.

\begin{figure}[t]
	\centering
	\includegraphics[width=0.9 \linewidth]{ 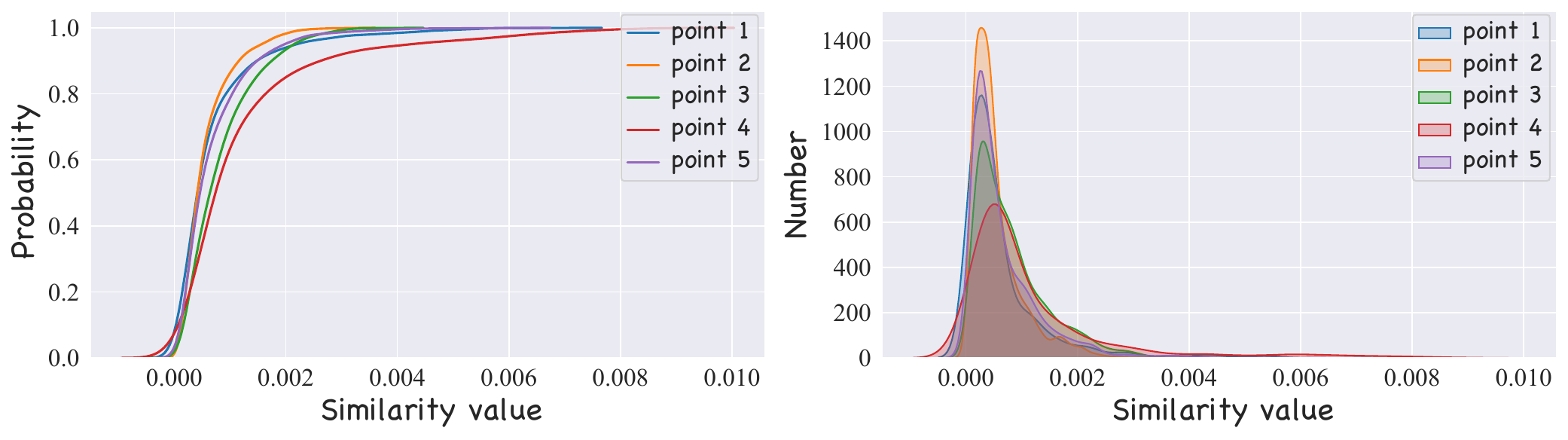}
	\caption{The cumulative distribution and 
	corresponding  histogram distribution of similarity values for 5 
	video points in $\boldsymbol{A}$ on the FCVID dataset~\cite{FCVID}, where $p=10$.}
	\label{fig:sim}
\end{figure}

% To avoid this dilemma, 
Different from~\cite{BTH}, we develop the Gaussian-adaptive similarity graph inspired by SSDH~\cite{SSDH}, 
which requires only one matrix to estimate more efficient sample relations.
Compared to SSDH, our novelty lies in building a graph based on each sample and mining hard negative samples.
% Because the neighbor distribution of each video is quite different, 
% it is too confusing to consider the overall distribution of all samples.
%受启发于
Specifically, we first investigate the cumulative distribution and corresponding histogram of similarity values for each video point in $\boldsymbol{A}$. 
For better visualization, we randomly select the similarity values corresponding to 5 video points on the FCVID dataset~\cite{FCVID}, and use kernel density estimation~\cite{KDE} to simulate the real distribution curve in Fig.~\ref{fig:sim}.
% we randomly pick 5 video points on the FCVID dataset~\cite{FCVID} and simulate the true curve using kernel density estimation~\cite{KDE} in Fig.~\ref{fig:sim}.
% By observing the cumulative distribution, we find that
Observing the cumulative distribution shows that the similarity values between most graph nodes are relatively small,  
while the histogram of the similarity value corresponding to each video point tends to a Gaussian distribution.
This shows from the real data that it is very noisy to directly treat all the similarity signals in $\boldsymbol{A}$ 
as positive samples.
% This shows from the real data that there are a lot of noisy supervision signals in $\boldsymbol{A}$, which will mislead the learning of hash functions if it is directly used.
To ensure high confidence in the supervision signal, we adaptively obtain positive samples for each video point.
For the video point $\boldsymbol{v}_{i}$, the mean and standard deviation of similarity values between the nodes
 can be expressed as $\mu_i$ and $\epsilon_{i}$, then we take the 
 positive sample estimator as $PT_i=\mu_i + \lambda_1 * \epsilon_i$.
Some metric learning works~\cite{zhao2018adversarial,milbich2020diva} show that hard negative samples are 
beneficial to model, so we add negative sample estimator $NT_i=\mu_i -  \lambda_2 * \epsilon_i$
to mine hard negative  samples for training.
In this way, the Gaussian-adaptive graph adjacency matrix can be expressed as:
\begin{equation}
	\hat{\boldsymbol{A}}_{i j}= \begin{cases}1, & \text { if } \boldsymbol{A}_{ij} \geq PT_i \\ -1, & \text { if } NT_i<\boldsymbol{A}_{ij}<\mu_i \\ 0, & \text {otherwise}  \end{cases}.
\end{equation}
%\begin{equation}
%\hat{\boldsymbol{A}}_{i j}= \begin{cases}1, & \text { if } \boldsymbol{A}_{ij} \geq PT_i \\ -1, & \text { if } NT_i<\boldsymbol{A}_{ij}<\mu_i \\ 0, & \text {otherwise}  \end{cases}.
%\end{equation}
To preserve 
the similarity graph relations mined in $\boldsymbol{\Omega}^{T}$  for binary codes,
we design a binary structure similarity loss:
\begin{equation}
	\mathcal{L}_{bsim}=\frac{1}{N} \sum_{\{i,j\} \in \mathcal{S}} |\hat{\boldsymbol{A}}_{i j}|(\hat{\boldsymbol{A}}_{i j}-\frac{1}{K} \boldsymbol{b}_{i} \boldsymbol{b}_{j}^{T})^{2},
\end{equation}
where $\mathcal{S}$ is the equal sampling strategy that samples positive or negative pairs with probability 0.5 based on $\hat{\boldsymbol{A}}$.
Finally, we can obtain discriminative  codes.

% , which samples positive or negative  pair with a probability of 0.5

% is the equal probability sampling strategy on positive 
% and negative video pairs based on $\hat{\boldsymbol{A}}$. 

Furthermore, some works~\cite{romero2014fitnets} argue that the middle layer of the teacher network can 
serve as a hint to the corresponding layer of the student network,
thereby improving the effect of semantic knowledge transfer.
Therefore, we consider aligning the visual embeddings between $\boldsymbol{\Omega}^{S}$and $\boldsymbol{\Omega}^{T}$.
Inspired by~\cite{tian2017deepcluster,greedy}, we design a visual embedding similarity loss:
\begin{equation}
\mathcal{L}_{tsim} = \frac{1}{N} \sum_{\{i,j\} \in \mathcal{S}} \left\|\overline{\boldsymbol{t}}_{i}-\boldsymbol{c}_{i 1}\right\|_{2}^{2} + \eta |\hat{\boldsymbol{A}}_{i j}|(1-\hat{\boldsymbol{A}}_{i j}) \big[\left\|\overline{\boldsymbol{t}}_{i}-\boldsymbol{c}_{i 1}\right\|_{2}^{2}-\left\|\overline{\boldsymbol{t}}_{i}-\boldsymbol{c}_{j 1}\right\|_{2}^{2}+ \beta \big]_+ ,
\end{equation}
where $\overline{\boldsymbol{t}}_i$ is the mean visual embedding of $i$-th video in $\boldsymbol{\Omega}^{S}$,
 $\boldsymbol{c}_{i1}$ or  $\boldsymbol{c}_{j1}$  is 1-NN nearest center of corresponding teacher visual embedding, $\eta$ controls the balance and $[x]_+$ means the hinge function $max(0,x)$, which makes $\overline{\boldsymbol{t}}_i$ closer to  $\boldsymbol{c}_{i1}$ than negative pair $\boldsymbol{c}_{j1}$ by a fixed margin $\beta=1.0$.

\subsection{Overall Learning}
The overall training objectives of $\boldsymbol{\Omega}^{T}$ and $\boldsymbol{\Omega}^{S}$ are as follows:
\begin{equation}
	\begin{gathered}
	\mathcal{L}_{\text {teacher }}=\mathcal{L}_{\text {recon }}, \\
	\mathcal{L}_{\text {student }}=\mathcal{L}_{\text {recon }}+\gamma_{1} \mathcal{L}_{\text {bsim }}+\gamma_{2} \mathcal{L}_{\text {tsim }},
	\end{gathered}
	\end{equation}
where $\gamma_1$ and $\gamma_2$  relatively weight the losses.
% The detailed training process can be found in the supplementary material.

% After DKPH is trained, we can obtain the binary codes for
% any out-of-sample videos as follows:

\section{Experimental Results}
\subsection{Datasets, Metrics and Implementation Details}
\textbf{Datasets.}
We run experiments on three popular video datasets.
\noindent \textbf{FCVID}~\cite{FCVID} is a web video dataset consisting of 91,223  YouTube videos annotated into 239 categories. 
It covers 
a wide range of topics, with the majority of them being real-world 
events such as “biking”, “making coffee” and “yoga”. 
The dataset is evenly split into training and testing partitions 
with 45,585 and 45,600 videos. 
We use the testing partition as the query set and retrieval database.
\noindent \textbf{ActivityNet}~\cite{activitynet} consists of 20K YouTube videos   annotated with 200 class descriptions.
  As the testing set labels are not publicly available, 
  the evaluation is performed on the validation set.
  Following~\cite{NPH}, we pick  9,722, 1,000 and 3,760 videos 
  as training set, query set and  retrieval database,
  respectively. 
\noindent \textbf{YFCC}~\cite{YFCC} is a massive dataset from the Yahoo Webscope program containing 0.8M videos. 
We randomly select 409,788 unlabeled videos for training 
and 101,256 labeled videos with 80 semantic concepts~\cite{xiao2010sun} for testing.
In these labeled videos, we sample 1000 videos as the query set
 and the remaining ones as retrieval database.

\noindent \textbf{Metrics.} 
We measure the retrieval performance with standard metrics in information retrieval,
including Mean Average Precision at top-k retrieved results (MAP@k) and
Precision-Recall (PR) curves.

\begin{figure}[t]
	\centering
	\includegraphics[width= 0.95\linewidth]{ 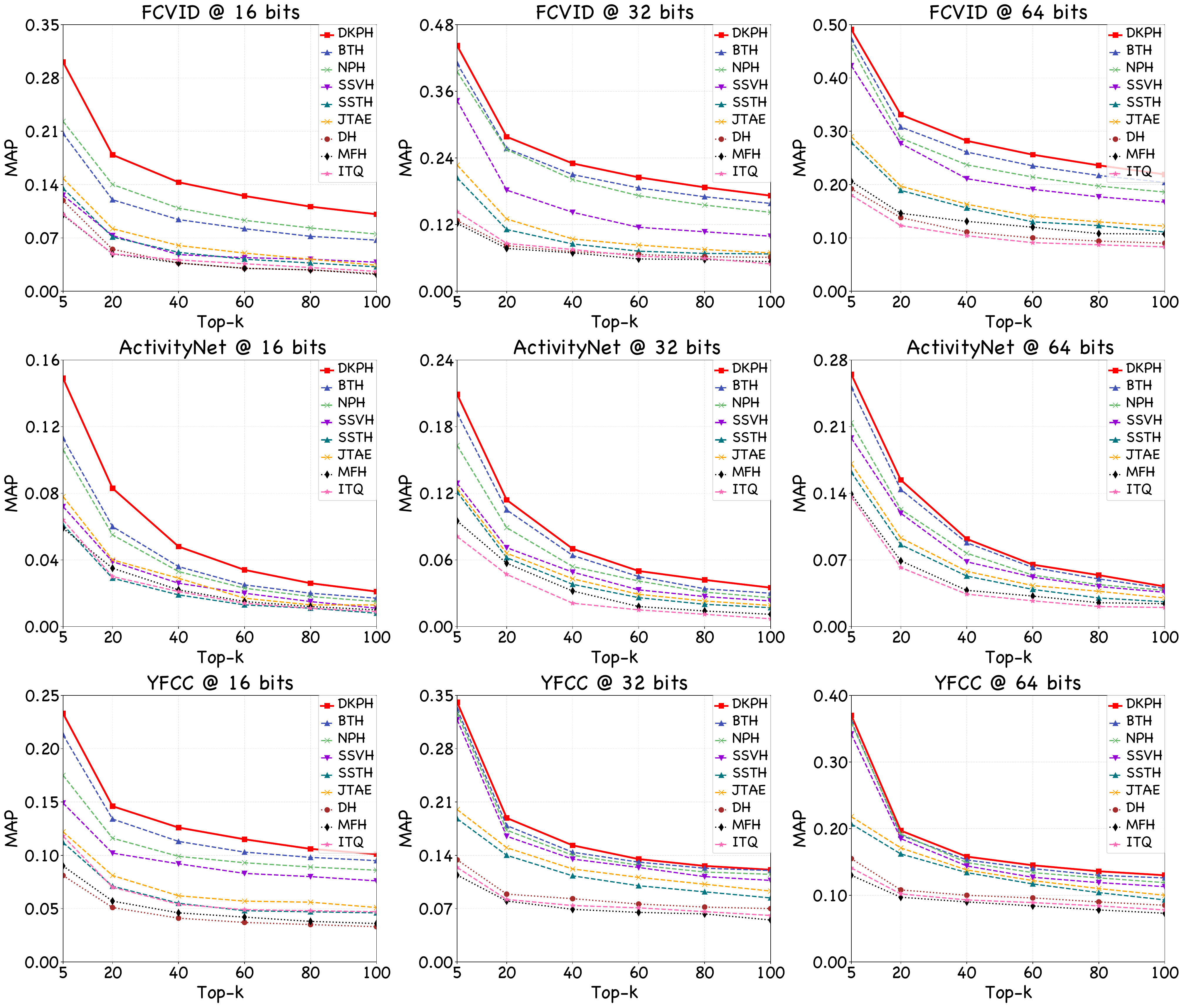}
	\caption{Performance comparison (\textit{w.r.t.} MAP@k) of DKPH and SOTA methods.} 
% 	the state-of-the-art unsupervised hashing methods on FCVID, ActivityNet and YFCC.}
	\label{fig:map}
\end{figure}

\begin{figure}[t]
	\centering
	\includegraphics[width= 0.95\linewidth]{ 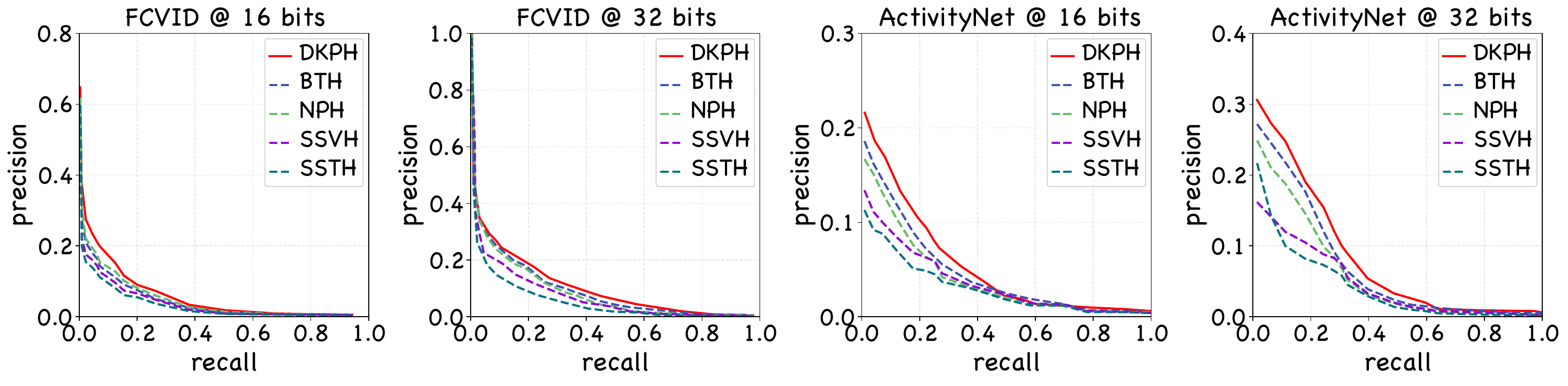}
	\caption{Performance comparison (\textit{w.r.t.} PR curves) of DKPH and SOTA methods.}
% 	the state-of-the-art unsupervised hashing methods on FCVID and ActivityNet.}
	\label{fig:pr}
\end{figure}

\noindent \textbf{Implementation Details.} 
% \subsection{Implementation Details} 
%    \noindent \textbf{Baselines.} 
Our experiments are based on the Pytorch framework~\cite{pytorch}.
% and executed on a single GTX TITAN X GPU with 12GB memory.
In the video encoding process, we uniformly sample 25 frames from each video and use VGG-16 pretrained on Imagenet~\cite{VGG} to extract frame-wise features.
To ensure fair comparison~\cite{BTH}, we use a single transformer block with a single attention head as the transformer encoder.
For the teacher model $\Omega^{T}$, we warm up 60, 300 and 200 epochs on FCVID, ActivityNet and YFCC, respectively.
% Then, we train the  student model $\Omega^{S}$ for 48 epochs .
Considering the trade-off of visual information compression loss and inter-frame correlations,
we set the dimension of the visual embeddings and binary codes to 256 and 128 in $\Omega^{T}$.
In the graph construction process, the number of clustering center $N_c$  
is set as 2,000, 1,000 and 2,000 on FCVID, ActivityNet and YFCC respectively. 
% $\alpha$ is adaptively set according to~\cite{AGH}.
% The dimension of visual embeddings $[\boldsymbol{t}_{i}^1, \cdots,\boldsymbol{t}_{i}^M]$  is 256.
We employ Adam optimizer to train the model  with a mini-batch size of 256 and train the student model $\Omega^{S}$ for 48 epochs, where the initial learning rate is $5 \times 10^{-4}$.
The default hyper-parameters setting is:
$\lambda_1 = 2, \lambda_2 = 1, \eta=0.1, \gamma_1=0.11, \gamma_2=0.9$.
In the testing phase, we only use the student model $\Omega^{S}$, where the lengths of binary codes are 16, 32 and 64.

\subsection{Comparisons with State-of-the-art (SOTA) Methods}
\vspace{-2pt}
% \noindent \textbf{Comparisons with State-of-the-art (SOTA) Methods.}
To prove the effectiveness of DKPH, we compare the retrieval performance with two image hashing methods:
   ITQ~\cite{ITQ}, DH~\cite{DH}, and six SOTA video hashing methods:
   MFH~\cite{MFH}, SSTH~\cite{SSTH}, 
   JTAE~\cite{JTAE}, SSVH~\cite{SSVH}, NPH~\cite{NPH}, and BTH~\cite{BTH}.
%   For image hashing methods, we directly use CNN frame features to 
%   supplement the experiment. 
   Fig.~\ref{fig:map} shows the MAP@K results on three datasets.
   Compared with SOTA methods, DKPH achieves the best results on three video datasets.
   Specifically, we obtain 0.7\%–8.6\% MAP@5 gains for various
   bits, which demonstrates the efficiency of DKPH.
   We owe the great advantage of DKPH over these two methods~\cite{BTH,NPH} to the full use of dual-stream structure and Gaussian-adaptive similarity graph.
   Note that the model performance gaps are larger at 16 bits, as we expected.
   Because the amount of information carried by the binary code is limited by the length.
  Therefore, in low-bit scenarios, the impact of task heterogeneity will be more serious, leading to inferior results from existing methods~\cite{BTH}.

% To further prove the effect on the  retrieval task, 
Furthermore, we examine DKPH with PR curves 
on FCVID and ActivityNet in Fig.~\ref{fig:pr}. 
DKPH delivers higher precision than SOTA methods at the same recall rate, and improves more significantly at low recall requirements.
This illustrates that the model is suitable for real-world video retrieval systems, as people tend to focus more on results with high accuracy rather than finding all similar results.
% This applies to real-world video retrieval systems, since people tend to 
% focus more on accurate results with high probability rather than finding all similar results.

\begin{table*}[t]

	\centering
	\renewcommand\arraystretch{0.85}
	\ttabbox{	\caption{ The impact of different frame feature encoders with or without 
		the dual-stream structure on FCVID. Some abbreviations: TF-Transformer, D-Dual-stream. }}{
	\scalebox{0.85}{
		\begin{tabular}{l|cccc|cccc|cccc}
			\hline
			\multirow{2}{*}{Method}   & \multicolumn{4}{c|}{16 bits}  & \multicolumn{4}{c|}{32 bits} & \multicolumn{4}{c}{64 bits}     \\  \cline{2-13} 
			& k=5 & k=20   & k=60 & k=100 & k=5 & k=20   & k=60 & k=100& k=5  & k=20  & k=60  & k=100  \\ \hline  	        
			CNN~\cite{yue2015beyond}           &0.229  &0.116   &0.080   &0.065
			&0.395   & 0.242 &0.172  &0.140  
			&0.460  &0.294  &0.207   & 0.171      \\   
			CNN~\cite{yue2015beyond}+D	&\textbf{0.273}  &\textbf{0.152}    & \textbf{0.106}   &\textbf{0.084}   
			&\textbf{0.407}  &\textbf{0.252}  & \textbf{0.175}  & \textbf{0.147} 
			&\textbf{0.464}   &\textbf{0.308}  &\textbf{0.223}   &\textbf{0.189}     \\   \hline
			LSTM~\cite{srivastava2015unsupervised}     &0.227  &0.114   &0.077   & 0.062 
			&0.393  &0.240  & 0.168  & 0.139 
			&0.457 &0.291  &0.210   & 0.174   \\ 	  	    
			LSTM~\cite{srivastava2015unsupervised}+D        &\textbf{0.272}   & \textbf{0.150}  &\textbf{0.104}   & \textbf{0.083} 
			 &  \textbf{0.404}& \textbf{0.248}  &\textbf{0.172} & \textbf{0.146}  
			 &  \textbf{0.462} & \textbf{0.301} &\textbf{0.224}   &\textbf{0.192}  \\  \hline
			TF~\cite{bert}	 & 0.235  & 0.122  & 0.083  &0.069  
			&0.421   &0.252  &0.172   &0.143  
			&0.477   &0.313  &0.238   &0.202   \\ 
			TF~\cite{bert}+D	&\textbf{0.297}  & \textbf{0.174} &\textbf{0.120} &\textbf{0.097}
		    &\textbf{0.441}    &\textbf{0.275} & \textbf{0.203}   &\textbf{0.171} 
		    &\textbf{0.494}    &\textbf{0.331}  & \textbf{0.255}   &\textbf{0.228}   \\ \hline 
		\end{tabular}
			} 
			\label{tab_ab1}	
			}

\end{table*}

\begin{table*}[t]
\centering
\renewcommand\arraystretch{0.9}
\ttabbox{\caption{ Contributions of different modules on FCVID.}}{
\scalebox{0.85}{

	\begin{tabular}{l|cccc|cccc|cccc}
		\hline
		\multirow{2}{*}{Method}   & \multicolumn{4}{c|}{16 bits}  & \multicolumn{4}{c|}{32 bits} & \multicolumn{4}{c}{64 bits}     \\  \cline{2-13} 
		& k=5 & k=20   & k=60 & k=100 & k=5 & k=20   & k=60 & k=100& k=5  & k=20  & k=60  & k=100  \\ \hline  	        
	DKPH+A	&0.202  & 0.119   &0.080  &0.062   & 0.399  &0.244  & 0.170 &0.139  &0.458  &0.294   &0.211 &0.180      \\   
	DKPH+M	&0.269 & 0.139  &0.091 & 0.075 &0.421   & 0.258 &0.176  &0.151  &0.484  & 0.321     & 0.243 & 0.204     \\ 
% 	\hline  
% 	DKPH-D	 & 0.223  & 0.130  & 0.088  &0.070  &0.417   &0.250  &0.172   &0.145  
% 			&0.476   &0.313  &0.231   &0.196   \\

	DKPH+T	 &0.228 &0.135 & 0.090 & 0.073 & 0.419   & 0.251  &0.172   &0.147 &0.463    &0.298   &0.218   & 0.181  \\ \hline
	DKPH-D	& 0.235  & 0.122  & 0.083  &0.069  
	&0.421   &0.252  &0.172   &0.143  
	&0.477   &0.313  &0.238   &0.202   \\ 
	DKPH-DR	 &0.276  &0.159  & 0.111  & 0.092  & 0.435    & 0.267   &0.189    &0.158  &0.487     &0.326    &0.245    & 0.218   \\ \hline
	DKPH-TS &0.158   &  0.088    & 0.067    & 0.051 &0.311   &0.123  &0.078  &0.058  &0.345  & 0.169     & 0.118&0.103        \\    
	DKPH-Lb	&0.174   & 0.093     & 0.072    &0.058  &0.322  & 0.142 & 0.080 &0.062  &0.366  & 0.187     &0.123 & 0.107       \\   
	DKPH-Lt &0.209   & 0.121  & 0.098    &0.080  &0.405   &0.239  &0.177  &0.141  &0.432  &0.276 &0.205 & 0.162     \\   \hline
	DKPH	&\textbf{0.297}  & \textbf{0.174} &\textbf{0.120} &\textbf{0.097}
	&\textbf{0.441}    &\textbf{0.275} & \textbf{0.203}   &\textbf{0.171} 
	&\textbf{0.494}    &\textbf{0.331}  & \textbf{0.255}   &\textbf{0.228}   \\ \hline 
	\end{tabular}  }
	\label{tab_ab2}
		}
\end{table*}

\begin{figure}[t]
	\centering
	\includegraphics[width= 0.97\linewidth]{ 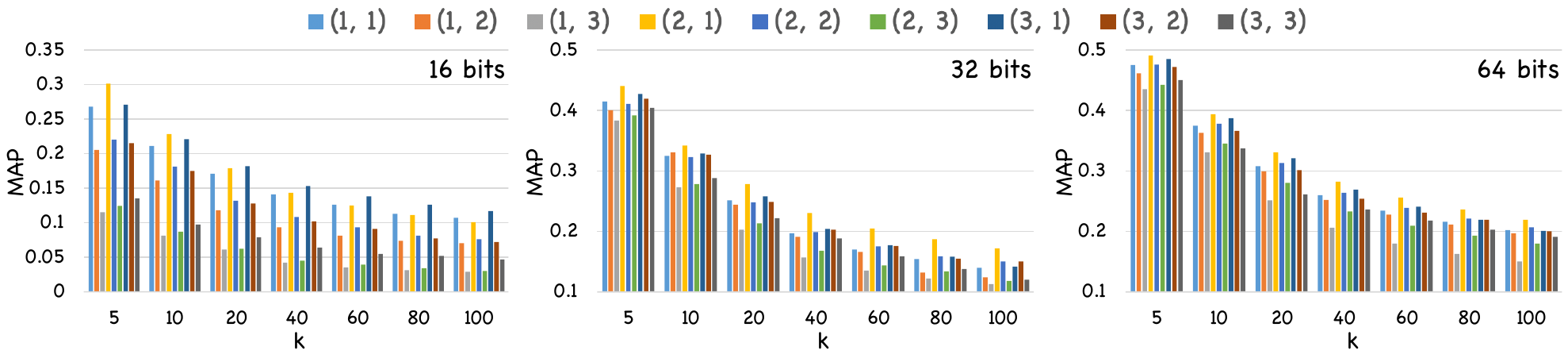}
	\caption{The MAP@k scores with various configurations about the positive estimator factor and the negative estimator factor $(\lambda_1, \lambda_2)$ on FCVID.}
	\label{fig:ablation_lambda}
\end{figure}

\subsection{Ablation study}
% \vspace{-4pt}
To provide further insight into DKPH, we conduct critical ablation studies.
%  to evaluate the effect of our model.

\noindent \textbf{Analysis of the dual-stream structure with different encoders.}DKPH employs a transformer encoder and a dual-stream structure to generate binary codes. 
Thus, we explore the impact of different frame feature encoders
(CNN~\cite{yue2015beyond}, LSTM~\cite{srivastava2015unsupervised}, and transformer~\cite{bert}) with or without the dual-stream structure in Table~\ref{tab_ab1}.
Specifically, based on the dual-stream structure, CNN, LSTM and transformer
obtain 0.4\%-6.2\% MAP@5 gains at different bits on FCVID.
The following advantages can be clearly observed:
(1) the dual-stream structure is a \textbf{general-purpose and important design} that consistently 
improves three encoders, especially at low bits;
(2) the transformer outperforms CNN and LSTM due to 
its strong ability to model long-term inter-frame correlations.

\noindent \textbf{Analysis of model components.} We compare DKPH with the following variations:
(1) DKPH+A. The Gaussian-adaptive graph adjacency matrix $\hat{\boldsymbol{A}}$ is replaced by $\boldsymbol{A}$; 
(2) DKPH+M. $\hat{\boldsymbol{A}}$ is replaced by  multiple matrices~\cite{BTH}; 
% (3) DKPH-D. We remove the dual-stream structure; 
(3) DKPH+T. We replace the dual-stream structure with twin bottlenecks designed for image hashing~\cite{shen2020auto}; 
(4) DKPH-D. The dual-stream structure is removed;
(5) DKPH-DR. Both the dual-stream structure and  $\mathcal{L}_{recon}$  in $\boldsymbol{\Omega}^{S}$ are removed;
(6) DKPH-TS. We remove  the  teacher-student distillation strategy and only use  $\mathcal{L}_{recon}$; 
(7) DKPH-Lb. We remove the binary structure similarity loss $\mathcal{L}_{bsim}$; 
(8) DKPH-Lt. We remove the visual embedding similarity loss $\mathcal{L}_{tsim}$. 
Table~\ref{tab_ab2} shows the performance of DKPH and its variations at different bits on FCVID,
and proves that each module significantly contributes to the final result.

We have the following observations.
% First, Table~\ref{tab_ab2} proves that each module significantly contributes to the final result.
First, reasonable mining of positive and hard negative pairs helps discriminate binary codes. 
 DKPH adaptively explores the similarity relations of each video point through the sample estimators.
However, in DKPH+M, the multiple matrices strategy consumes more time and space resources, and requires careful adjustment of matrix parameters, which cannot achieve optimal results.
Second, twin bottlenecks (DKPH+T) are still difficult to replace the dual-stream structure designed for video hashing. 
There are two reasons: (1) the mechanism of twin bottlenecks is to learn better reconstructed images to feedback binary codes, which cannot exhibit the advantages of video information decomposition; 
(2) twin bottlenecks generate frame-level binary codes, resulting in quantization errors during testing.
%and quantization errors will still occur during testing;
% (3) twin bottlenecks introduce the graph convolution structure with more complex computation.
Third, DKPH-D and DKPH-DR explore the effects of task heterogeneity, which suggests a conflict between $\mathcal{L}_{recon}$ and similarity learning in existing methods~\cite{SSVH,BTH}.
We decouple the tasks, which allows binary codes to retain useful information and avoids the conflict.
Fourth,  results in DKPH-Lb yield an excessive drop due to the lack of similarity guidance, 
where $\mathcal{L}_{bsim}$ is the core loss of  information decomposition.

\noindent \textbf{Hyperparameter analysis.} We investigate various configurations about the positive and negative estimator factors $(\lambda_1, \lambda_2)$, as shown in Fig.~\ref{fig:ablation_lambda}.
From this experiment, we find that as $\lambda_1$ grows (\textit{i.e.,} $PT_i$ grows), the
performance increases at first and reaches the best results,
then decreases as a whole.
A small $PT_i$ may cause the model to be trained on more noisy signals, 
, while a large
$PT_i$ may not fully exploit the underlying positive similarity relations.
Moreover, $\lambda_2$ has a greater impact on model performance than $\lambda_1$, reflecting the vital contribution of hard negative samples to the model.

 \begin{table}[t]
	\centering
	\renewcommand\arraystretch{1.05}
		\ttabbox{\caption{ Cross-dataset MAP@20 results when training 
   		on FCVID and test on YFCC at 64 bits. The blue number indicates the 
   		performance drop compared with training and testing both on YFCC (black number).}}{
	\scalebox{0.9}{
		\begin{tabular}{l|c| c| c| c |c}
			\hline
			{Method \qquad}   &SSTH~\cite{SSTH} &SSVH~\cite{SSVH} & NPH~\cite{NPH} & BTH~\cite{BTH} & DKPH\\ \hline  	        
			MAP@20  &0.155 \  ( \textcolor{blue}{-6.3\%} )  &0.173\  ( \textcolor{blue}{-7.8\%} )  &0.180\ ( \textcolor{blue}{-6.0\%} )    &0.191\ ( \textcolor{blue}{-5.7\%} )  &   0.199\ ( \textcolor{blue}{-2.8\%} )       \\   \hline  
		\end{tabular}
			}  
			\label{tab_cross}	
			}

	\end{table}

\begin{table*}[t]
%	\floatsetup{floatrowsep=qquad}  
	\begin{floatrow}
		\begin{minipage}{0.33\linewidth}
			\centering
			\renewcommand\arraystretch{0.9}
			\ttabbox{}{%
		\begin{tabular}{l|c c c }
	\hline
	{Method}   &k=5 &k=20 & k=60   \\ \hline  	  
 Only $\boldsymbol{l}$  &0.098   &0.077  &0.065   \\   \hline       
 Only $\boldsymbol{b}$ 	&\textbf{0.297}     & \textbf{0.174}       &\textbf{0.120}     \\   \hline  
	
\end{tabular}
\caption{The MAP@k results of latent features $\boldsymbol{l}$ and binary codes $\boldsymbol{b}$ at 16 bits. }
}
		\end{minipage}
		
		\begin{minipage}{0.33\linewidth}
			\centering
			\renewcommand\arraystretch{0.9}
\floatsetup{floatrowsep=qquad,captionskip=10 pt }
%			\begin{floatrow}
			\ttabbox{}{%
				\begin{tabular}{l|l}
						\hline
	{Method \qquad}   & {Error \qquad}   \\ \hline  	  
    DKPH  &\textbf{0.4849}\\   \hline   
	$\text{DKPH}_f$ &0.9527 \\   \hline      
				\end{tabular}
			\caption{The reconstruction errors in a  category. }
			}
		\end{minipage}
		
	\begin{minipage}{0.33\linewidth}
			\centering
			\renewcommand\arraystretch{0.9}
\floatsetup{floatrowsep=qquad,captionskip=10 pt }
%			\begin{floatrow}
			\ttabbox{}{%
				\begin{tabular}{l|l}
						\hline
	{Method \qquad }  &{Error \qquad}  \\ \hline  	  
DKPH  &\textbf{0.5586}\\   \hline  
Remove $\boldsymbol{b}$   &0.5615\\   \hline  
Remove $\boldsymbol{l}$	&0.9769 \\   \hline      
	\end{tabular}
			\caption{The reconstruction errors in test set. }
			}
		\end{minipage}
	\end{floatrow}
\end{table*}

\subsection{Further Analysis}
	 \begin{figure}[t]
		\centering
		\includegraphics[width= 0.9\linewidth]{ 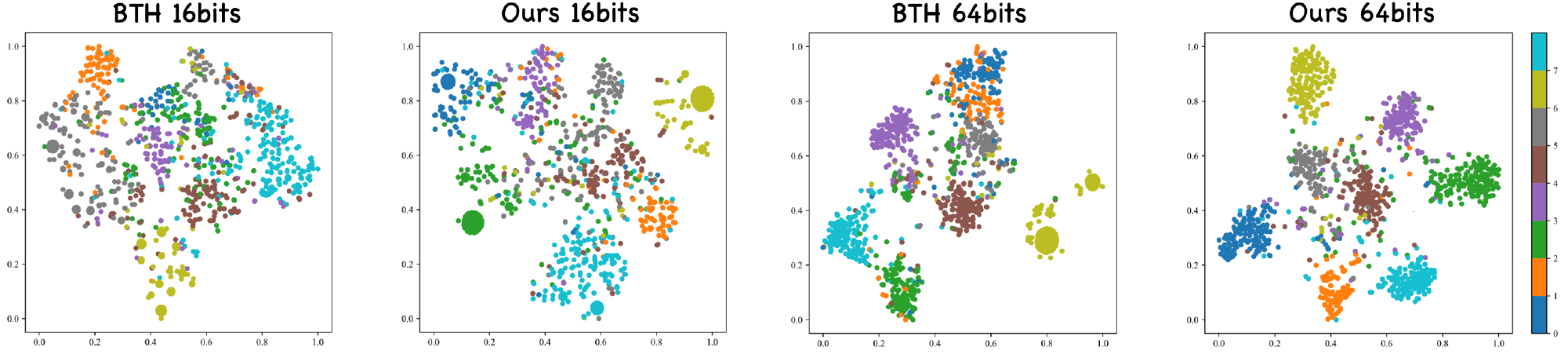}
		\caption{t-SNE visualizations~\cite{tsne} of BTH and DKPH.
		 Videos are randomly sampled on FCVID database, and samples with different labels are marked with different colors.}
		\label{fig:tsne}
	\end{figure}

\noindent \textbf{Cross-dataset evaluation comparisons.}
To investigate  the generalization of DKPH for cross-dataset retrieval,
we train  various methods on FCVID and test on YFCC in Table~\ref{tab_cross}, which shows MAP@20 results for cross-dataset retrieval at 64 bits.
DKPH can not only achieve SOTA in the single-dataset setting, 
but also the performance drop (-2.8\%) is the lowest in the cross-dataset setting.
This may be because binary codes focus more on semantic concepts rather than the underlying reconstruction information, which ensures  good transferability and generalization of DKPH when retrieving unknown datasets.

\noindent \textbf{Information decomposition analysis.}
Table 4 shows the MAP@k results of latent features $\boldsymbol{l}$ and binary codes $\boldsymbol{b}$ at 16 bits on FCVID.
Results of Only $\boldsymbol{l}$ are much lower than those of Only $\boldsymbol{b}$, which indicate that $\boldsymbol{l}$ may not have enough semantics to support the retrieval task.
Next, we examine the effect of dual-stream features for reconstruction at 16 bits on FCVID. 
In Table 5, we randomly input a category of test videos and then calculate the mean square error between $[\boldsymbol{x}_{i}^1, \cdots,\boldsymbol{x}_{i}^M]$ and $[\tilde{\boldsymbol{x}}_{i}^1, \cdots,\tilde{\boldsymbol{x}}_{i}^M]$.
% the reconstructed frame features and the original frame features.
When we replace $\boldsymbol{l}$ with fixed values (\textit{i.e.,}the mean of latent features), 
the reconstruction error in $\text{DKPH}_f$ increases by 96.5\%.
In Table 6, we directly remove $\boldsymbol{b}$ or  $\boldsymbol{l}$ for reconstruction and calculate errors in all test videos. 
Removing $\boldsymbol{b}$, the error increases by 0.52\%, while Removing $\boldsymbol{l}$ increases the error by 74.9\%.
Table 5 and 6 prove that $\boldsymbol{l}$, rather than $\boldsymbol{b}$, contains sufficient essential information (dynamic changes) for reconstruction.

\noindent \textbf{Qualitative results.}
%  Top-5 retrieval results are illustrated in Fig.~\ref{fig:retrieval}.
% It is difficult for SSVH to distinguish different semantics, 
% our model can still search videos in the same class, which proves
% the discrimination of binary codes with the knowledge preservation strategy.
Fig.~\ref{fig:tsne} shows the t-SNE visualization~\cite{tsne} of binary codes learned by BTH and DKPH. 
To facilitate the observation, we randomly sample 8 categories of videos twice 
on 16 bits and 64 bits, respectively, to obtain binary codes.
 At 16 bits, there is a clear distinction between most categories in our model.
 In particular, t-SNE embeddings of DKPH in some categories (\textit{e.g.,} 0, 2, 6) 
 can be mapped onto a small circle.
This proves that DKPH pays more attention to the learning of global semantics and binary codes of a category are almost very close in Hamming space, so the phenomenon of t-SNE embedding aggregation occurs.
 At 64 bits, t-SNE embeddings of our model in different categories are well separated, which proves the good discriminativeness.

 \section{Conclusion}
We propose a novel unsupervised video hashing framework, DKPH, to tackle the task heterogeneity problem.
Firstly, we design the dual-stream structure to decompose video information, which disentangles the semantic extraction from reconstruction constraint.
Then, a Gaussian-adaptive similarity graph is developed to explore the semantic similarity knowledge between samples.
With the help of this knowledge, the hash layer in the dual-stream structure can further generate discriminative semantic binary codes.
In this paper, we hope not only to present insights into the importance of information decomposition but also to facilitate future work that advances video hashing by solving design flaws rather than mostly trial and error.

% Teacher-student optimization further guides the hash layer to leverage the semantic similarity knowledge to generate discriminative binary codes.
% Teacher-student optimization. Knowledge distillation further helps the hash layer to exploit the potential of semantic similarity learning to generate discriminative binary codes.

~\\
\noindent \textbf{Acknowledgements.}
This work is supported by the National Nature Science Foundation of China (62121002, 62022076, U1936210), 
the Fundamental Research Funds for the Central Universities under Grant WK3480000011, the Youth Innovation Promotion Association Chinese Academy of Sciences (Y2021122).
We acknowledge the support of GPU cluster built by MCC Lab of Information Science and Technology Institution, USTC.

\clearpage

\bibliographystyle{splncs04}
\bibliography{my_ref}
\end{document}